\documentclass[journal]{IEEEtran}
\usepackage{amsmath,amsfonts}
\usepackage{algorithmic}
\usepackage{algorithm}
\usepackage{array}
\usepackage[caption=false,font=normalsize,labelfont=sf,textfont=sf]{subfig}
\usepackage{textcomp}
\usepackage{stfloats}
\usepackage{url}
\usepackage{verbatim}
\usepackage{graphicx}
\usepackage{cite}
\usepackage{hyperref}
\usepackage{fontawesome5}
\usepackage[framemethod=tikz]{mdframed}
\definecolor{mycolor}{rgb}{0.122, 0.435, 0.698}
\usepackage{tcolorbox}
\newtcolorbox{notebox}{
	colback=red!5!white,
	colframe=red!100!black,
	coltext=red!100!black,arc=0pt,halign=center,}

\DeclareMathOperator*{\argmin}{arg\,min}

\usepackage{tikz}

\usepackage{amsthm}
\newtheorem{prop}{Proposition}
\newtheorem{lem}[prop]{Lemma}
\newtheorem{cor}[prop]{Corollary}
\newtheorem{thm}[prop]{Theorem}

\theoremstyle{definition}
\newtheorem{defi}[prop]{Definition}

\newcommand{\R}{\mathbb{R}}
\newcommand{\N}{\mathbb{N}}

\newcommand{\LR}{\mathcal{L}_\R}
\newcommand{\sumdata}{\sum_{i=1}^{n}}

\begin{document}

\title{From Features to Graphs: Exploring Graph Structures and Pairwise Interactions via GNNs\thanks{This work has been submitted to the IEEE for possible publication. Copyright may be transferred without notice, after which this version may no longer be accessible.}
}


 \author{Phaphontee Yamchote, Saw Nay Htet Win, Chainarong Amornbunchornvej,  Thanapon Noraset
         }


%
%

\markboth{Journal of \LaTeX\ Class Files,~Vol.~14, No.~8, August~2021}%
{Shell \MakeLowercase{\textit{et al.}}: A Sample Article Using IEEEtran.cls for IEEE Journals}


\maketitle

\begin{abstract}
	Feature interaction is crucial in predictive machine learning models, as it captures the relationships between features that influence model performance. In this work, we focus on pairwise interactions and investigate their importance in constructing feature graphs for Graph Neural Networks (GNNs). We leverage existing GNN models and tools to explore the relationship between feature graph structures and their effectiveness in modeling interactions. Through experiments on synthesized datasets, we uncover that edges between interacting features are important for enabling GNNs to model feature interactions effectively. We also observe that including non-interaction edges can act as noise, degrading model performance. Furthermore, we provide theoretical support for sparse feature graph selection using the Minimum Description Length (MDL) principle. We prove that feature graphs retaining only necessary interaction edges yield a more efficient and interpretable representation than complete graphs, aligning with Occam's Razor.
	Our findings offer both theoretical insights and practical guidelines for designing feature graphs that improve the performance and interpretability of GNN models.
\end{abstract}

\begin{IEEEkeywords}
Feature interaction, Graph neural networks
\end{IEEEkeywords}

\section{Introduction}\label{section:intro}

One of the key challenges in predictive modeling is the behavior of data whose two or more independent variables exhibit an interaction effect on their dependent variable -- we refer to this as \textbf{Feature Interaction} \cite{molnar2020interpretable}. The interaction refers to conjunctive operations, such as multiplication, rather than treating features in isolation or through simple linear combinations. For example, if we have a dataset generated by the equation $y = x_1x_2 + x_3$, then we say that $x_1$ and $x_2$ interact with each other in the model of $y$. Applications of feature interaction modeling span a wide range of domains, particularly the click-through rate (CTR) problem, where they play a critical role in improving predictions \cite{liu2022gain, lyu2023optimizing, wang2024click}.
For example, the user's age and product type may interact differently on various platforms.
Younger users may prefer to buy fashion products (young and fashionable), while older users might lean towards formal products (old and formal).




\begin{figure}[t]
	\centering
	\includegraphics[width=1\linewidth]{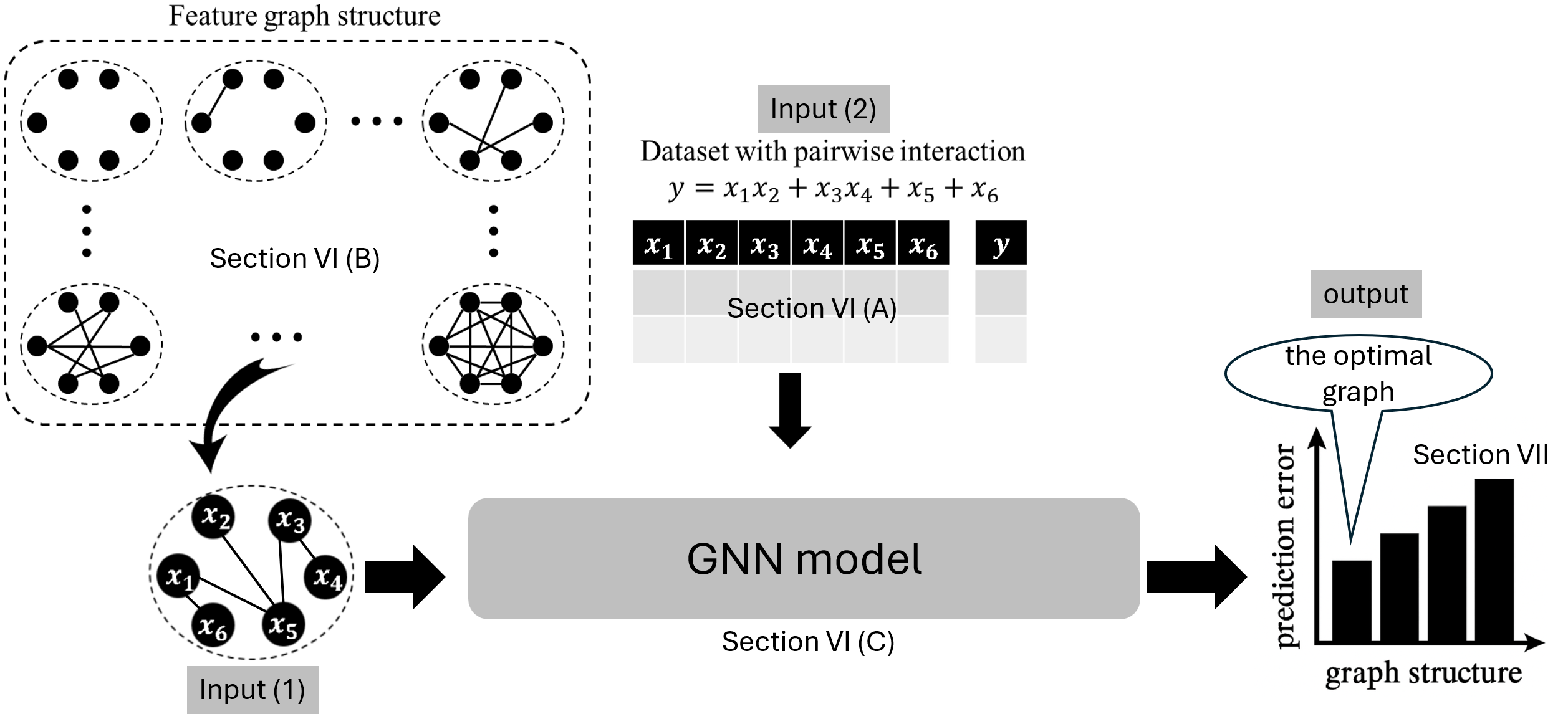}
	\caption{High-level figure to show the overall objective of our study: Given a dataset consisting of pairwise feature interaction, which structure of feature graph should we use as an input data graph structure for the given GNN model? Specifically, which graph structure can give the highest performance to GNNs? And what are the dependencies between these features that the optimal graph structure tells us?
	}
	\label{fig:firstfigure}
\end{figure}

Modeling these interactions between variables has been a long-standing research problem in machine learning.
Traditionally, these interactions were captured through manual feature engineering, where practitioners explicitly crafted cross-features based on domain knowledge \cite{DeepFM}.
Powered by the theory for the partial dependence decomposition, Friedman's H statistic \cite{Hstat} is a method to learn the dependence between features.
While effective in some cases, manual approaches are labor-intensive, unscalable, and prone to missing complex or subtle interactions \cite{molnar2020interpretable}.
To address these limitations, models such as the Cross Field Network (CFN) \cite{wang2022deepcross} were developed to automatically learn bounded-degree feature interactions, particularly in click-through rate prediction tasks.
Recently, graph neural networks have emerged as a powerful framework for modeling interactions through graph-structured representations.
In this setting, feature graphs, where nodes represent features, play an important role in GNN-based models for modeling interactions \cite{kim2022explicit}. Figure \ref{fig:firstfigure} illustrates examples of feature graphs when a set of features in a dataset is given.

While feature graphs are more expressive in representing feature interactions with edges between nodes, a major challenge arises when we do not know the interactions in advance.
Thus, most methods rely on searching or learning from a complete graph, which includes every possible interaction, i.e., every pair of features, to avoid missing any potentially important interaction \cite{fignn}.
For example, Table2Graph \cite{Table2Graph} formulates the construction of a unified feature graph as a reinforcement learning problem, where the graph's edge weights are optimized using the reinforce algorithm.
In addition, Fi-GNN \cite{fignn} takes advantage of the robust expressive capabilities of feature graphs to represent the multi-field features in a feature graph structure where different fields can interact through edges.
Although these show the potential applications of feature graphs, a key theoretical question remains: {\em what characteristics should a feature graph have to optimally model interactions?} Understanding the relationship between feature graph structures and their ability to model feature interactions is essential to building efficient and effective algorithms for manually constructing feature graphs when interactions are known or learning feature graphs when interactions are unknown.

In this work, instead of proposing another feature graph algorithm, we attempt to discover the insights related to the feature graph structures and feature interactions, focusing on pairwise interactions. Our key contributions are as follows.
\begin{enumerate}
	\item \textbf{Theoretical Foundation for Sparse Feature Graphs:}
	We establish a theoretical foundation for sparse feature graphs using the Minimum Description Length (MDL) principle. By framing the feature graph selection problem in the context of MDL, we prove that retaining only necessary interaction edges minimizes the description length, making it a better choice than using complete graphs. (Section \ref{sec:theoResult})  
	\item \textbf{Insights into Feature Graph Structures:}
	We provide insights into the structures of feature graphs pertaining the given feature interactions instead of using complete graphs to be used as input graphs for GNN models.
	Our experiment result shows that the edges corresponding to the pair of interacting features are significant and must remain in the feature graphs to maintain the prediction performance. (Section \ref{sec:insight-results})
	\item \textbf{Empirical Validation on Real-World Datasets:}
	We empirically show that, on real-world datasets, the feature graphs constructed by the pairwise interaction can improve the prediction performance compared with baselines, including tree-based algorithms and linear-feature-based algorithms. (Section \ref{sec:caseStudy}) 

\end{enumerate}

\section{Preliminary}\label{sec:prelim}
\begin{table}[!h]
	\centering
	\begin{tabular}{|c|l|}
		\hline
		Notation   & Description \\ \hline
		$\R$       & the set of real numbers   \\ \hline
		$\R^d$     & the set of d-dimensional real-valued vector   \\ \hline
		$x$        & a real feature value  \\ \hline
		$\vec{x}$  & a real-valued feature vector   \\ \hline
		$G$        & a graph   \\ \hline
		$V$ & a set of nodes of a graph $G$   \\ \hline
		$E$ & a set of edges of a graph $G$   \\ \hline
		$A$ & an adjacent matrix of a graph $G$   \\ \hline
		$X$ & an embedding matrix of a graph $G$   \\ \hline
		$\mathcal{N}(v)$ & a set of nodes adjacent to the node $v$ in a graph $G$   \\ \hline
		$\mathcal{G}(d)$ & a set of feature graphs of $d$ nodes $v_1, \dots, v_d$    \\ \hline
		$v$ & a node of a graph    \\ \hline
		$e_{uv}$ & an edge from a node $u$ to a node $v$    \\ \hline
		$\vec{h}_i^{l}$ & an embedding of a node $v_i$ in the layer $l$  \\ \hline
		$\vec{e}_{uv}$ & an embedding of an edge $e_{uv}$  \\ \hline
		$\vec{m}_i^l$ & a message aggregated from $\mathcal{N}(v_i)$ in the layer $l$   \\ \hline
		$\mathcal{L}(H)$ & description length for a model $H$  \\ \hline
		$\mathcal{L}(D|H)$ & description length for a dataset $D$ encoded\\
		& by a model $H$  \\ \hline
		$\mathcal{L}(D)$ & minimum description length for a dataset $D$  \\ \hline
		$\LR(x)$ & the number of bits of  $x\in\R$ by some encoding  \\ \hline
		\end{tabular}
		\caption{All related terminology}
\end{table}

\paragraph*{Feature Interactions} Feature interaction refers to the operation between two or more features concerning the output variable. Linear combinations or additive models represent zero-order interactions, in other words, no interaction \cite{fignn}. 
For example, in a ground truth model, $y = 5x_1 + x_2x_3 - 7\sin(x_4 - 3x_5x_6)$, we observe interactions between $x_2$ and $x_3$, as well as between $x_4, x_5$ and $x_6$ with respect to $y$.

\paragraph*{Feature Graphs and Graph Neural Networks}
For graphs, we denote a tuple $G = (V, E, A,X)$ of nodes $V$, a set of edges $E$, an adjacency matrix $A$ and an embedding matrix $X$.
For any node $ v \in V $, we denote by $ \mathcal{N}(v) $ the set of nodes adjacent to $ v $, called the neighborhood of the node $ v $.
If only the structure of a graph is considered, we may use only $G(V,E)$. Since we represent the interaction between features by graph edges, we model instances with the graphs whose nodes correspond to features of a dataset.
According to Fi-GNN \cite{fignn}, this graph is called a \textbf{feature graph}. (see the example in Figure \ref{fig:experiment})
Mathematically, given a dataset of features $x_1,\dots, x_d$, a feature graph structure is a graph $G(V,E)$ where $V = \{v_1,\dots, v_d\}$.


There are many tools for processing graphs. GNN models have gained significant attention due to their effectiveness in handling complex, non-Euclidean data.
GNNs are a category of neural models specifically designed to operate on graph data structures. In these structures, the order of nodes should not influence the computation of predicted values, a property referred to as ``permutation invariance." Various approaches are used to implement GNNs, but one of the most prominent paradigms is based on ``message passing."

In our context, message passing can be described through the following equations:
\begin{align}
	\vec{m}_{i}^{(k)} &= \bigoplus_{j\in \mathcal{N}(i)} \phi^{(k)} \left(\vec{h}_{i}^{(k-1)},\vec{h}_{j}^{(k-1)}, \vec{e}_{j,i}\right)\label{message}
	\\
	\vec{h}_{i}^{(k)} &= \gamma^{(k)} \left(\vec{h}_{i}^{(k-1)}, \vec{m}_{i}^{(k-1)}\right)\label{updating}
\end{align}
where $\bigoplus$ represents a differentiable, permutation-invariant function, such as summation, averaging, or maximum operation, while $\gamma$ and $\phi$ denote differentiable functions, often implemented as neural network layers.
In Equation \ref{updating}, the model updates the node information using both existing information (from the previous layer) and information shared among neighboring nodes (messages), which is computed as in Equation \ref{message}. The latter equation pertains to the aggregation of messages from neighboring nodes.

Message passing layers are not the sole focus in GNN models; pooling layers also play a crucial role, particularly in tasks like graph classification. Pooling involves summarizing local information in a graph, such as node embeddings, to derive a global representation of the entire graph. There are various methods for graph pooling, including sum pooling and mean pooling \cite{SelfAttentionGP}. 

\paragraph*{Minimum Description Length (MDL)}
Our problem can be framed as the selection of models, specifically the graph representations.
MDL \cite{MDL, Rissanen1978ModelingBS} is a paradigm of model selection that attempts to find the model that minimizes the number of bits representing information.
This information may include the complexity of models and the performance score of the model on given data.

In our work, we use MDL to study a model whose complexity is minimized while maintaining the prediction's well-performed accuracy.
It is the two-stage code description approach \cite{Grnwald2007TheMD}. 
Specifically, it encodes the data $D$ with the description length $\mathcal{L}_\mathcal{H}(D)$ by first encoding a hypothesis $H$ from the set of given hypotheses $\mathcal{H}$ followed by encoding $D$ with the help of $H$.
The objective is to minimize the total description length as follows:
$$
\mathcal{L}_{\mathcal{H}}(D) := \min_{H\in\mathcal{H}}(\mathcal{L}(H)+ \mathcal{L}(D|H)).
$$
In other words, $\mathcal{L}(H)$ includes any descriptions related to the model, such as coefficients in linear regression or the number of edges in the representation of the feature graph (as used in this work).
The latter term, $\mathcal{L}(D|H)$, typically includes the code length of the prediction errors and the code length of the data.

\section{Related Work}\label{sec:related}

\subsection{Feature Interaction Modeling}

Feature interactions in terms of nonadditive terms play a crucial role in model-based predictive modeling.
Many works attempted to develop methods to indicate the interaction.
Beginning with the statistical-based measure called Friedman's H statistic \cite{Hstat} involving the theory of partial dependency decomposition.
However, it is computationally expensive and is dependent on predictive models.

From the machine learning perspective, many models were proposed to model feature interactions, especially variants of factorization models inspired by FM \cite{FM}.
FM relies on the inner product between embedding the feature-value, motivated by the factorization of matrices.
Several variants of FM have been proposed to fill the gaps.
FFM \cite{FFM} takes into account the embedding of feature field (i.e. columns in a table of data).
AFM \cite{AFM} considers the weight of the interaction by adding attention coefficients to its model.

The success of deep nets in various domains motivates researchers to use them in modeling interactions.
For example, Factorization Machine supported Neural Network (FNN) \cite{FNN} proposed a deep net architecture that can automatically learn effective patterns from categorical feature interactions in CTR tasks.
In addition, DeepFM \cite{DeepFM} proposed a kind of similar idea but included a special deep net layer that performs the FM task.

The attention mechanisms \cite{Bahdanau} are a method designed to model the importance of different components.
In AFM, the attention mechanism was applied because the authors believe that not all interactions should be treated equally likely with respect to prediction values.
Attention allows the model to weigh feature interactions differently based on their importance.
After introducing the attention mechanism to interaction tasks, many works use attention to model the interaction of features based on the factorization machine framework \cite{Sarkar2022DualAH, Cheng2019AdaptiveFN,Wang2020AdnFMAA,Wen2020NeuralAM,Li2021GraphFMGF}.

One crucial challenge of interactions in traditional machine learning models is the explainability of which features have interactions, although some of them outperform in terms of prediction performance.
So we need an explicit representation that can correspond to pairwise interaction, where a graph is a representation that can fill this gap.
In addition, GNN is a tool that can be used to learn information from graphs.

\subsection{Graph Neural Networks for Feature Interactions}
GNNs are becoming more interesting for use in feature interaction problems.
Since feature interactions can be seen as relationships between features, most of the works rely on feature graphs for each individual instance to represent the interactions via the aggregation of representations from neighboring nodes.

Motivated by the click-through rate (CTR) prediction, which is believed to influence the rate of clicking on a product, such as age and gender, many works attempted to model this prediction using GNN applied to feature graphs.
To the best of our knowledge, Fi-GNN \cite{fignn} is the first work to use GNN to model the interaction of features in the feature graph for the prediction of CTR.
It applied the field-aware embedding layer to compute the latent representation of nodes of features before feeding the feature graph with the computed representations into a stack of message passing layers.

After the introduction of Fi-GNN, many subsequent works have built on this idea.
Cross-GCN \cite{Feng2020CrossGCNEG} used a simple graph convolutional network with cross-feature transformation.
GraphFM \cite{GraphFM} seamlessly combined the idea of FM and GNN using the neural matrix factorization \cite{NeuralMF} based function to estimate the edge weight. It also adopts the attentional aggregation strategy to compute feature representation.
Table2Graph \cite{Table2Graph} applied attention mechanisms to compute the probability adjacency matrix instead of edge weight. It also used the reinforcement learning policy to capture key feature interactions by sampling the edges within feature graphs.

All mentioned works chronically improve the methods of feature interaction learning.
However, no work provides information about the effect of the construction of feature graphs on the learning capacity.
Most works utilize attention mechanisms or edge weights to let models learn graph structures via these parameters.

\section{Problem Formulation: Pairwise Interaction Feature Graph Problem} \label{sec:problemformulation}

A crucial open research question in graph representation for data without explicit connections is that {\em What is a suitable graph structure?} While current research highlighted the potential of feature graphs, a feature graph that is too dense, for example, a complete graph, may lead to learning issues, as shown and discussed in previous work \cite{fignn, Kipf2016SemiSupervisedCW, Rong2019DropEdgeTD} and in Section \ref{subsec:complete-graph}.
Thus, the community's common goal is to find an optimal feature graph structure for any particular prediction task.
Consequently, we can formulate the ``feature-graph-finding'' problem as follows:
\begin{equation*}
	G^* = \argmin_{G\in\mathcal{G}(n)} \ell(\vec{y}, \text{GNN}(X; G)),
\end{equation*}
where $\text{GNN}(\ \cdot\ ;G)$ is a given GNN model trained for the feature graph $G$, and $\ell$ is an arbitrary but fixed loss function for the prediction task. The above task is not traceable as there are exponentially many possible graph structures ($\mathcal{O}(2^{n^2})$.) Many works attempted to solve this problem using soft learnable interactions \cite{fignn,GraphFM} or hard interactions through reinforcement learning \cite{Table2Graph}. However, they did not provide insight into the characteristics of a suitable feature graph structure related to the interaction between features.

Consequently, instead of solving the ``feature-graph-finding'' problem, this work aims to study the relationship between feature graph structures and their performance in modeling feature interactions.
Specifically, we would like to theoretically show that (1) feature graphs correspond to feature interactions and (2) keeping only edges between interacting feature nodes is sufficient under MDL assumptions.

\section{Theoretical Results}\label{sec:theoResult}
There are two questions that we should address.
The first is ``Is the space of the functions and the space of feature graphs a correspondence?''
It is significant because we aim to guarantee that we can use some graph to represent any of the given functions and vice versa.
The second question is that of ``Why do we need sparse graphs rather than complete graphs?''
Here, we consider the second problem to be the problem of model selection according to the concept of MDL.
Proofs of results in this section are in the Appendix.

\subsection{Correspondence Problem}\label{subsubsec:corresQues}

It is worth considering the correspondence between expressions and graph structures.
Even though we primarily focus on pairwise interactions, in this correspondence, we go towards the arbitrary number of variables interacting, not only pairwise.
For convenience of stating, we discuss only the multiplicative interactions, which is a special case of the non-additive interaction.
Throughout this, let us call the kind of expression disjoint multilinear interaction which is defined as follows:

\begin{defi}[Disjoint Multilinear Interaction Expression]
	The predictive model $y(\vec{x})$ is said to be a \textbf{disjoint multilinear interaction expression} (DMIE) if it can be expressed as 
	\begin{equation}
		y(\vec{x}) = \sum_{i=1}^n \prod_{j=1}^{m_i} x_{i_j}
	\end{equation}
	where $x_{i_j}$ and $x_{k_l}$ are distinct variables when $i\neq k$ or $j\neq l$.
	We denote by $\mathcal{DM}(n)$ the set of all DMIE expressions of $n$ variables.
\end{defi}

Multiple graphs can infer the same expression, considering the perspective of connected components. For example, the expression $x_0 x_1 x_2 + x_3 x_4 + x_5 + x_6$ corresponds to the graphs $E_1 = \{\{0,1\},\{0,2\},\{1,2\},\{3,4\}\}$ and also $E_2 = \{\{0,1\},\{1,2\},\{3,4\}\}$. Consequently, we use equivalence classes of graphs (based on connectedness) instead of individual graphs. To this end, we define an equivalence class of feature graphs.

\begin{defi}
	We define a binary relation $\sim$ in $\mathcal{G}(n)$ as follows: for any graph $G_1,G_2 \in \mathcal{G}(n)$, $G_1\sim G_2$ if for any $i, j\in N_n$, they are reachable in $G_1$ if and only if they are reachable in $G_2$.
	
	Obviously, $\sim$ is an equivalence relation to $\mathcal{G}(n)$.
	We then denote the set of equivalence classes from $(\mathcal{G}(n),\sim)$ by $\mathfrak{G}(n)$:
	\begin{equation}
		\mathfrak{G}(n) := \left\{ \left[G\right]_\sim : G \in \mathcal{G}(n) \right\}
	\end{equation}
\end{defi}

As a result, we can prove the correspondence, as presented in Theorem \ref{mainThm}.

\begin{thm}\label{mainThm}
	$\mathcal{DM}(n)$ one-to-one corresponds to $\mathfrak{G}(n)$.
\end{thm}

\subsection{MDL for graph representation of pairwise interaction}\label{subsec:MDL}

This problem can be considered as a model selection problem for selection of representations.
We adopt the idea of Minimum Description Length (MDL) to shape the problem in the manner of selection of graph representation.
Generally speaking, a selected feature graph should not be so dense that it captures an uninformative characteristic.
On the other hand, it should not be such a light graph that the informative edges to learn interactions are omitted.
In this work, we focus on the pairwise interaction case, which is formalized below.

Given a graph $G=(V,E)$ whose nodes are of degree at most 1, 
A function containing pairwise interaction induced from $G$ is defined as
\begin{equation}\label{eq:pairwisefunction}
	f_G(x) = \sum_{\deg(i)=0} c_i x_i + \sum_{\{i,j\}\in E} c_{ij}x_i x_j
\end{equation}
Note that we represent pairwise interaction by bilinear interaction for convenience in describing by mathematical notation.
Therefore, the problem of selecting the graph for this function $f_G$ is framed as Problem \ref{problem:MDLInteractionCapture}.

\begin{algorithm}
	\floatname{algorithm}{Problem}
	\caption{MDL Pairwise Interaction Graph Problem}\label{problem:MDLInteractionCapture}
	\textbf{Input:} A dataset $\mathcal{S} = \{(x_i,y_i)\}_{i=1}^n$, and a pre-configured GNN model $\mathcal{F}:\mathfrak{G}\times\R^d  \to \R$\\
	\textbf{Output:} The edge set $E^*$ for an input feature graph $G^*(V,E^*) \in \mathfrak{G}$ so that
	\begin{equation}
		G^* = \argmin_{G\in\mathfrak{G}}{\mathcal{L}(f_G) + \mathcal{L}(S|f_G)}
	\end{equation}
\end{algorithm}
\noindent In this problem, we denote $\mathcal{L}(S,f_G) = \mathcal{L}(f_G) + \mathcal{L}(S|f_G) $, where
\begin{align}
	\mathcal{L}(f_G) &= \LR (|E(G)|) + \sum_{\deg(i)=0}\LR(c_i) + \sum_{\{i,j\}\in E(G)}\LR(c_{ij})\\
	\mathcal{L}(S|f_G) &= \sumdata\LR(x_i) + \sumdata \LR(y_i - f_G(x_i))
\end{align}
\noindent We denote by $\LR:\R\to\N$ a function of the number of bits $\LR(x)$ we need to encode a real number $x$.
In the next part, we will show that the description length of the optimal graph $G^*$ is bounded by the lengths of both the complete graph $K$ and the null graph $N$.

\paragraph*{Assumptions}
	Our consequent result in this paper is based on these assumptions about $\LR$:
	\begin{enumerate}
		\item $\forall x_1,x_2\in\R,\LR (x_1+x_2) \leq \LR(x_1) + \LR(x_2)$
		\item $ \forall x_1,x_2\in\R, |x_1|\leq |x_2|\iff\LR(x_1)\leq\LR(x_2) $
		\item $
		\exists D >0\forall x,y\in \R,|x-y| \leq 1 \Rightarrow |\LR(x) - \LR(y)| \leq D
		$
	\end{enumerate}

		We assume that the functions that generate $Y$ from $X$ in a dataset, denoted by $S_{G^*} := \{(X_i,Y_i)\}_{i=1}^n$, are induced by some feature graph $G^*$ as Equation \ref{eq:pairwisefunction}, i.e. $$Y = f_{G^*}(X) + \mathcal{E}$$ where $\mathcal{E}$ is a noise random variable from some truncated distribution, i.e. there exists $\epsilon^*>0$ with $|\mathcal{E}|\leq\epsilon^*$.
		
		We also denote $\hat{f}_{G}$ a function $f_G$ corresponding to a feature graph $G$ whose parameters are approximated by $S_{G^*}$.
		Moreover, we assume that the parameters of actual additive terms can be learned well (by some learning algorithm).
		Specifically, if the term $c_ix_i$ exists in the ground truth function, then the parameters $d_i$ of the feature graph can be learned so that $\LR(d_i - c_i)\approx 0$. It is similar for the interaction term $c_{ij}x_ix_j$, the parameter $d_{ij}$ can be learned so that $\LR(d_{ij} - c_{ij})\approx 0$.
		
		We also assume further that if the features $a$ and $b$ interact with each other in the expression of ground truth, then $\LR(\hat{d}_a) + \LR(\hat{d}_b) >> \LR(\hat{c}_{ab})$ where $\hat{d}_a$ and $\hat{d}_b$ are the coefficients learned by the expression that these features are additive and $\hat{c}_{ab}$ are those learned by the expression consisting of the interaction term between $x_a$ and $x_b$.
		So, it is safe to assume that $\LR(\hat{d}_a) + \LR(\hat{d}_b) - \LR(\hat{c}_{ab}) \geq D$ which is a uniform bound of difference between two real numbers that is not further than 1.
		On the other hand, we also assume that if $a$ and $b$ do not interact with each other, then $\LR(\hat{c}_{ab}) >> \LR(\hat{d}_a) + \LR(\hat{d}_b)$.
		
		For the description length of error term $\sumdata \LR(y_i - f_G(x_i))$, we need the assumption that fitting a pairwise interaction term $c_{ab}x_ax_b$ by the additive terms $c_ax_a + c_bx_b$ always yields dramatic length with respect to the noise term $\epsilon_i$ so that
		$$
		\LR(c_{ab}x_ax_b -(\hat{c}_ax_a + \hat{c}_bx_b) + \epsilon_i) >> \LR(\epsilon_i).
		$$
		Similarly, we also need a similar assumption to fit the additive terms $c_ax_a + c_bx_b$ by the interaction term $c_{ab}x_ax_b$ as follows:
		$$
		\LR((c_ax_a + c_bx_b) - \hat{d}_{ab}x_ax_b + \epsilon_i) >> \LR(\epsilon_i).
		$$

First, we begin with the most intuitive result motivated by our empirical result that we must keep the edges of interactions and we should prune non-interaction edges out in the desired feature graph.

\begin{lem}\label{lemma:error_term}
	If $\{a,b \}\in E(G^*)$, then we have $$\sumdata \LR(y_i - f_{G^*\setminus\{a,b\}}(x_i)) \geq \sumdata \LR(y_i - f_{G^*}(x_i)).$$
	If $\{a,b \}\notin E(G^*)$, then $$\sumdata \LR(y_i - f_{G^*\cup\{a,b\}}(x_i)) \geq \sumdata \LR(y_i - f_{G^*}(x_i)).$$
\end{lem}

\begin{prop} \label{prop:add-itr-drop-nonitr-from-good}
	If $\{a,b \}\in E(G^*)$, then we have $ \mathcal{L}(S_{G^*},f_{G^*\setminus\{a,b\}})\geq \mathcal{L}(S_{G^*},f_{G^*})$. On the other hand, if $\{a,b \}\notin E(G^*)$, then $ \mathcal{L}(S,f_{G^*\cup\{a,b\}})\geq \mathcal{L}(S_{G^*},f_{G^*})$.
\end{prop}

By Lemma \ref{lemma:error_term}, it is easy to show that:
\begin{cor}\label{cor:MDLerrorComplete}
	Let $K$ be the complete feature graph of the features of $S_{G^*}$.	Then
	$$
	\sumdata \LR(y_i - \hat{f}_K(x_i)) \geq \sumdata \LR(y_i - \hat{f}_{G^*}(x_i))
	$$ 
	Consequently, we obtain the desired result.
	
	$$\mathcal{L}(S_{G^*},f_{G^*}) \leq \mathcal{L}(S_{G^*},f_K).$$
\end{cor}

Not only for the complete graph, but also for the graph without edges we aim a similar result.

\begin{cor}\label{lem:MDLerrorNull}
	Let $N$ be the feature graph without edges.
	Then
	$$
	\sumdata \LR(y_i - \hat{f}_N(x_i)) \geq \sumdata \LR(y_i - \hat{f}_{G^*}(x_i))
	$$
	Hence, we also have $\mathcal{L}(S_{G^*},f_{G^*}) \leq \mathcal{L}(S_{G^*},f_N)$.
\end{cor}
Finally, we get the result that in aspect of MDL, complete graph and null graph are worse than any graph containing all interaction edges stated as follows:
\begin{thm} \label{theorem:complete-null-bad}
	For any feature graph $G$ containing all interaction edges, we have
	$$
	\mathcal{L}(S_{G^*},f_{G}) \leq \mathcal{L}(S_{G^*},f_K)
	$$
	$$
	\mathcal{L}(S_{G^*},f_{G}) \leq  \mathcal{L}(S_{G^*},f_N)
	$$
\end{thm}

\paragraph*{Implication}

The theoretical results shown in Section \ref{subsubsec:corresQues} are just to ensure us that we can confide in learning feature graph structures to infer an expression containing feature interactions.
In other words, given any function containing feature interactions of at most one occurrence for each feature, there always exist feature graph structures that can infer backward to the function and vice versa.
Therefore, to find such graph structures, we need some insight into graph construction related to feature interactions.

Consequently, we reformulate these observations into more general arguments.
Generally speaking from empirical results in Section \ref{subsec:itr-vs-nonitr}, it is better to maintain all the edges of the interaction in the feature graph to maximize the learning interaction in a given dataset.
Also, it is better to prune non-interaction edges out as much as possible to avoid fitting noise information.
Finally, these results are supported by Proposition \ref{prop:add-itr-drop-nonitr-from-good}.

In addition, we also show the empirical result about using complete feature graphs to learn from larger datasets in Subsection \ref{subsec:complete-graph}.
Although the result in Subsection \ref{subsec:itr-vs-nonitr} cannot infer the limitation of complete graph obviously,
the result in Subsection \ref{subsec:complete-graph} reveals that the complete feature graphs are worse when datasets are larger in feature size.
Finally, we can show that, in the manner of MDL which is a framework of model selection, the complete feature graph is worse than any feature graphs containing all interaction edges.
This tells us that it is not a good idea to use the complete feature graph to learn from datasets.
Not only learning hidden noise, but it also faces computation and memory issues.
The following section will focus on the empirical results in more detail.

\section{Methodology}\label{sec:methodology}

\begin{figure*}[t!]
	\centering
	\includegraphics[width=\linewidth]{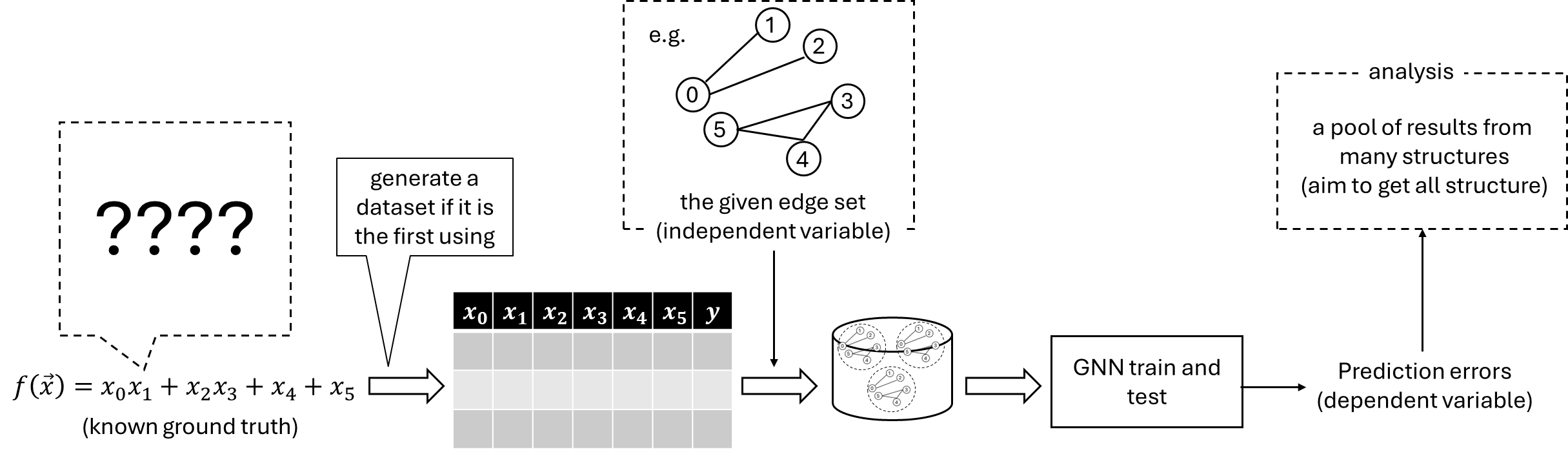}
	\caption{Experimental procedure: data generation based on a ground truth function and variation of feature graph structures as an input.}
	\label{fig:experiment}
\end{figure*}

Our experiment provides empirical evidence regarding the relationship between feature graph structures and dataset pairwise interactions. We train and evaluate several feature graph structures for each dataset and report the results.
However, due to the combinatorial number of possible feature graph structures and the stochastic nature of the learning algorithm, we adopt a strategy of generating random edge sets to explore various graph configurations, as illustrated in Figure \ref{fig:experiment}. 

\subsection{Datasets}
Real-world datasets rarely provide information about interactions between features, which disables our analysis.
To this end, we use synthesized datasets to control interaction characteristics in datasets such as the number of interaction pairs of features, the strength of interactions, or the kinds of interactions. We assume that there is only one mode of interaction per feature graph.
If real-world datasets had many interaction modes, we could apply the weighted ensemble feature graphs proposed by Li et al. \cite{Li2021GraphFMGF}

Construction of the simulated dataset (the first arrow in Figure \ref{fig:experiment}) can be formally described as follows.
Let $f:\R^d\to\R$ be a hidden ground truth function containing pairwise interaction terms. A synthetic dataset of $n$ instances with respect to $f$ if $\{(\vec{X_i},f(\vec{X_i}) + 0.1\mathcal{E}_i)\}_{i=1}^n$ where $\vec{X_i}\sim\mathcal{N}(0,1)$ and $\mathcal{E}_i\sim\mathcal{N}(0,\text{var}(f(\vec{X_i})))$. We varied the number of features and interaction pairs to aid our analyses.
Note that for each function $f$, we generated one set of data and used it for all variations of feature graphs.

\subsection{Feature graphs} 
For a given function $y=f(x_1,x_2,\dots, x_d)$ of $d$ features, a set of feature graphs is constructed by randomly assigning edges between feature nodes (see Figure \ref{fig:experiment} after the second arrow).
In other words, given a dataset of $d$ features and $n$ samples $\{ (\vec{x}_i, y_i) \}_{i=1}^n$ and an edge set $E$ for a graph of $d$ nodes, the feature graph dataset of edge set $E$ is a graph dataset
$
\left\{(G(V,E,\text{concat}(\vec{x},X)),y_i)\right\}_{i=1}^n,
$
where $X$ is a learnable embedding for a feature graph for nodes.
The initial node embedding matrix is denoted by
\begin{equation}
    H^{(0)}=\text{concat}(\vec{x},X). \label{eq:initEmb}
\end{equation}

\subsection{GNN Architectures}
\begin{figure*}[h]
	\centering
	\includegraphics[width=0.65\linewidth]{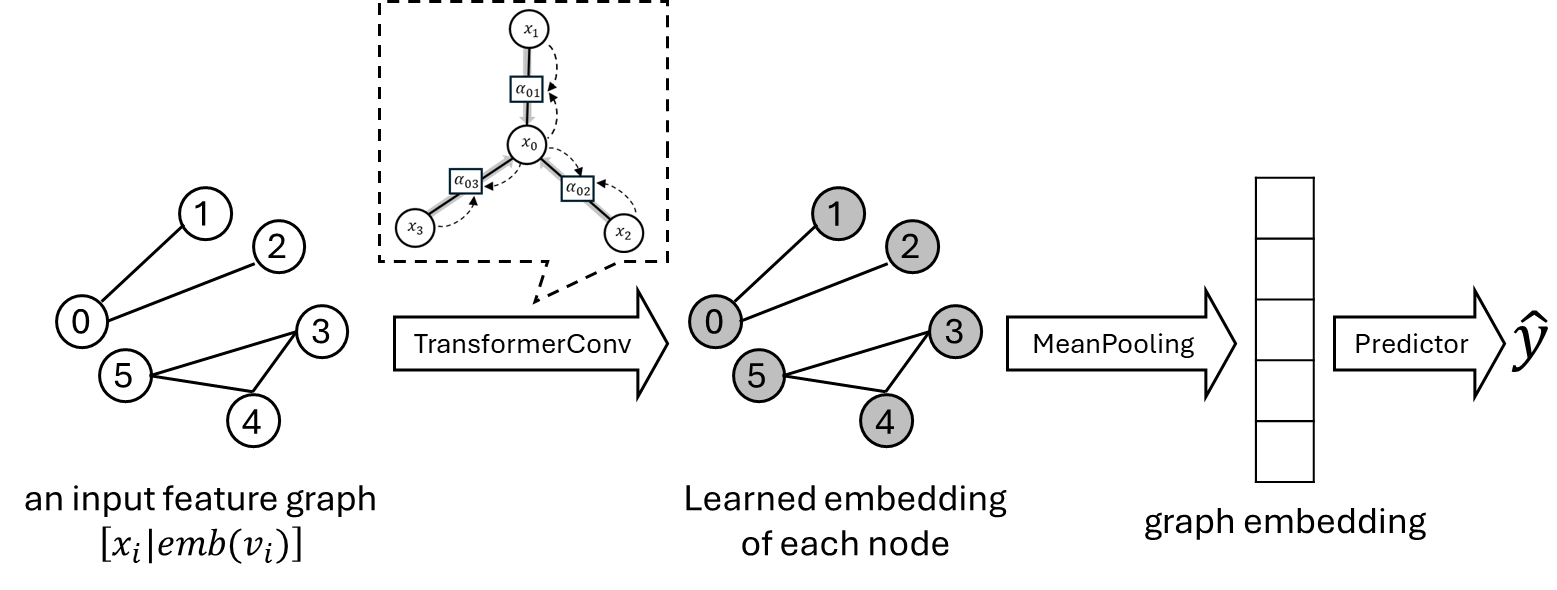}
	\caption{(will be fixed for used notation) Detail of the GNN model: from feature graph construction to output prediction.}
	\label{fig:overview}
\end{figure*}

Our GNN model utilizes transformer convolution layers (TransformerConv \cite{TransformerConv}) for the message passing function, incorporating attention mechanisms to capture the strength of pairwise interactions, as depicted in Figure \ref{fig:overview}. Then, a mean pooling layer averages the embedding of all nodes for the prediction layer.

As Fi-GNN does, this work adopts the idea from Fi-GNN, but with a simpler architecture.
We use TransformerConv \cite{TransformerConv}, which is a message-passing layer based on the attention mechanism, whose feature interaction learning capability is shown by many works.
For the computation, we start from the initial node embedding $H^{(0)}$ defined in Equation \ref{eq:initEmb}.
Then, a feature graph with the node embedding matrix is fed to TransformerConv with BatchNorm:
\begin{equation} \label{eq:transformerConv}
	\begin{split}
    	Z^{(l)} &= \text{TransformerConv}(H^{(l-1)},E),\\
    	H^{(l)} &= \text{ReLU}(\text{BatchNorm}(Z^{(l)})),
	\end{split}
\end{equation}
where $H^{(l)}$ is an intermediate node embedding matrix at the layer $l$.
After applying all message passing layers, we calculate the average pooling of node embeddings from the last GNN layer, and we use this vector as a graph embedding, which is fed into the projection layer to get a prediction:
\begin{align} 
    	\vec{p} &= \frac{1}{n}\sum_{i}H^{(L)}[i],\label{eq:pooling}\\
    	\hat{y}  &= \vec{w}\cdot \vec{p} + b,\label{eq:finalProjection}
\end{align}
where $H^{(L)}$ is a node embedding matrix from the last GNN layer and $H^{(L)}[i]$ is the row $i$ of the matrix which corresponds to the node embedding of the node $i$ and $b$ is a biased term.
Equation \ref{eq:pooling} is the calculation of the mean pooling layer, and Equation \ref{eq:finalProjection} is the final projection to calculate the prediction value.

Equation \ref{eq:transformerConv} is a bit different from the original work \cite{TransformerConv}, which uses LayerNorm after aggregation of messages, similar to the original transformer in NLP.
In the experiment's first phase, we found that the GNN model with LayerNorm cannot capture interaction behavior while BatchNorm can.

In training the GNN model, we adopt a usual supervised learning approach, where the loss function measures the difference between the predicted value, $\hat{y}$, and the target value, $y$.
Specifically, we use the mean squared error (MSE) loss, which is defined as
$$
\mathcal{L}_{\text{MSE}} = \frac{1}{N} \sum_{i=1}^{N} (\hat{y}_i - y_i)^2,
$$
where $N$ is the number of samples.

\section{Experiments and Results}\label{sec:insight-results}

Results shown in this section are from the dataset generated by the equation $f(\vec{x}) = \sum_{i=1}^{p} x_{2i-1}x_{2i} + \sum_{i=1}^{q}x_{2p+i} + \epsilon$.
We present the experiment results in the following insights:
\begin{enumerate}
	\item The efficient structures of feature graphs (Section \ref{subsec:itr-vs-nonitr})
	\item The dependency between edges and interaction pairs (Section \ref{subsec:remove-itr})
	\item The relationship between multi-hops and number of message passing layers (Section \ref{subsec:reach})
	\item Scalability of efficient feature graphs (Section \ref{subsec:complete-graph})
\end{enumerate}
Since Section \ref{subsec:itr-vs-nonitr} - \ref{subsec:reach} consider the variants of graph structures w.r.t. the given dataset, we show only results from an equation containing 2 interaction terms and 2 unary terms, i.e. $p =2, q=2$.
In Section \ref{subsec:complete-graph}, $p$ is varied while $q=2$.
\subsection{Experimental Setting}
\paragraph*{Synthetic dataset generating}

Each synthetic dataset contains 10,000 samples generated by distribution described in Methodology section (Section \ref{sec:methodology}).
In every randomization, we use seed number $n$ for feature $x_n$ and seed number 0 for noise term.
We split it into first 7,000 rows for training and next 3,000 rows for testing sets.

\paragraph*{Random sampling of feature graph structures}
Ideally, we want results of all possible feature graph structures.
However, there are up to $\mathcal{O}(2^{n^2})$ feature graphs for a dataset of $n$ features.
This leads us to the combinatorial issue,
so we set up an experiment by sampling in random stratified by (1) the number of edges, (2) the number of interaction edges, (3) the number of non-interaction edges, and (4) the number of hops between interacting nodes.
Since we also want to study about subgraphs with respect to existence of interaction edges (for example, edges $\{0,1\}$ and $\{2,3\}$ for the dataset generated by $y=x_0x_1 + x_2x_3 + x_4 + x_5$), we want these collections (1) graphs having both edges, (2) graphs having $\{0,1\}$ but not $\{2,3\}$ (3) vice versa and (4) graphs having no these edges. To have full collections to compare, once any feature graph is sampled, all other three graphs are also sampled.

\paragraph*{Model setting and training}
We also vary the number of message passing layers from 1 layer to 3 layers where all models' hyperparameters are set to have the equally likely number of parameters. In all setting, we use the 16-dimensional node embedding $X$. The hidden size in TransformerConv $H^{(i)}$ for 1, 2 and 3 MP layers is 26, 20, and 16, respectively.
In training of models, we use Adam optimizer with automatically schedule of learning rate.
%

\subsection{Interaction Edges vs. Non-interaction Edges} \label{subsec:itr-vs-nonitr}
We are first interested in the existence of interaction edges and that of non-interaction edges.
Intuitively, the existence of interaction edges should be significant for learning the interaction between features by attention-based GNNs since they allow paasing of the message between their associated nodes to model interactions.

\begin{figure}[h]
	\centering
	\includegraphics[width=\linewidth]{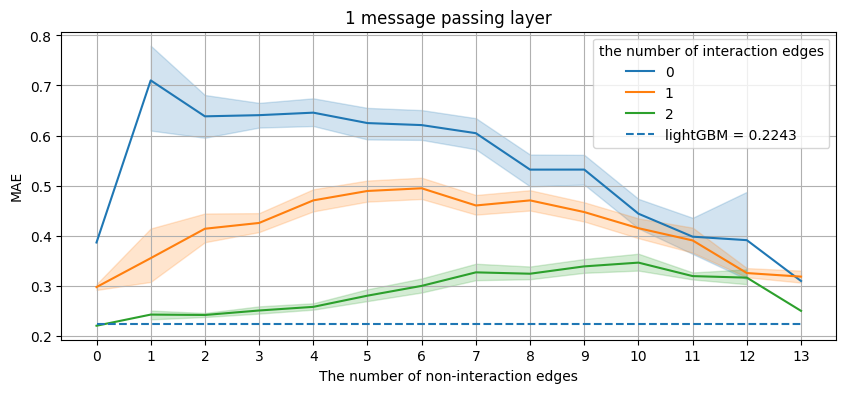}
	\includegraphics[width=\linewidth]{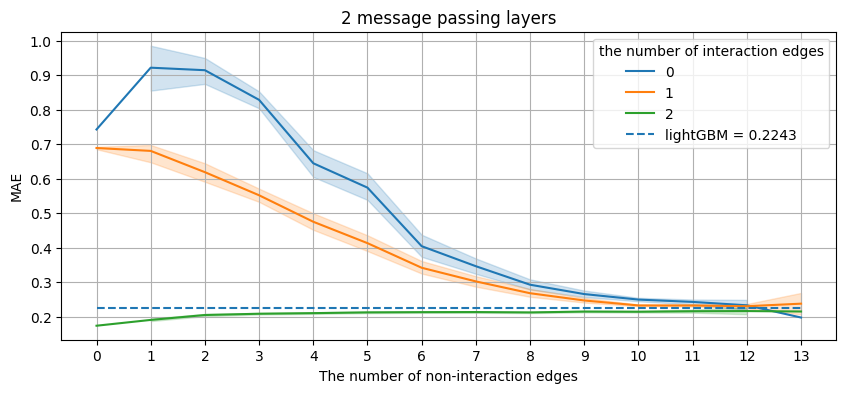}
	\includegraphics[width=\linewidth]{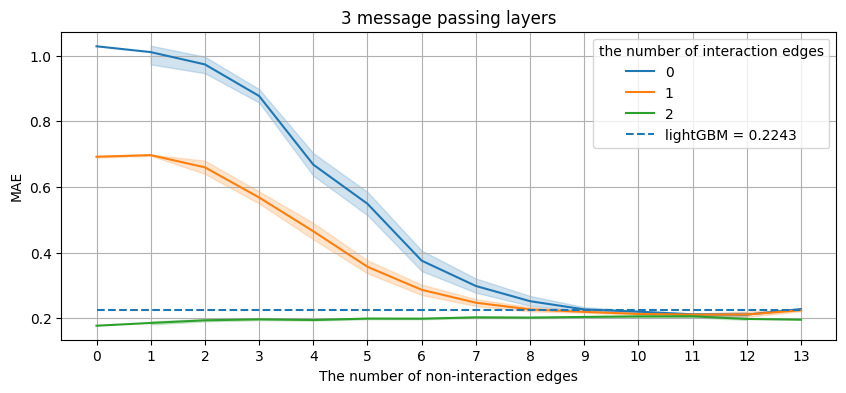}
	\caption{the average of MAE (y-axis) in prediction against the number of non-interaction edges (the edges corresponding to the nodes of features not interacting in equation, e.g. the edge $\{0,4\}$ or the edge $\{4,5\}$) in x-axis where different lines in the same figure indicated by color are for the number of interaction edges (there are two possible interaction edges in this equation: $\{0,1\}$ and $\{2,3\}$)}
	\label{fig:result1-1}
\end{figure}


Figure \ref{fig:result1-1} illustrates the average MAE in predicting each number of edges without interaction along the $x$ axis.
Each line corresponds to the number of interaction edges.
We see that among the graphs that have the same number of non-interaction, feature graphs containing all interaction edges give significantly lower errors than the graphs missing some interaction edges give for all of the number of non-interaction edges.
Especially, we also see that even though graphs with all interaction edges contain any number of other noise edges, they still give good performance compared to the entire line of the graphs missing some interaction edges. 
From this result, we may infer that the interaction edges are the most important in input feature graphs to learn a dataset containing pairwise interaction.
Although we do not know how to construct feature graphs used by GNN models for the task that we believe that there must be interaction terms, at least the interaction edges (if we know) should be kept.

This result also tells us that when the feature graph can maintain all interaction edges, the increasing number of non-interaction edges leads to poorer performance.
It seems that the non-interaction edges make a model capture non-underlying information in a dataset and interfere with the learning of pairwise interactions of interaction edges.

However, if we consider the results of the 0-interaction-edge and 1-interaction-edge graphs, we notice a different trend from the 2-interaction-edge graphs.
It starts with the worst result when there are no edges in a graph.
In this case, the more noninteraction edges it has, the better the performance.
One hypothesis of us at first sight is that the more number of edges (even though they are not interaction edges), the higher chance that interaction nodes can pass messages in GNNs' computation via multi-hop connectivity rather than direct interaction edges.
This leads us to consider the connectivity of interacting nodes rather than the direct edges between them.

\subsection{Removing Interaction Edges from a Graph} \label{subsec:remove-itr}
\begin{figure}[h!]
	\centering
	\includegraphics[width=0.7\linewidth]{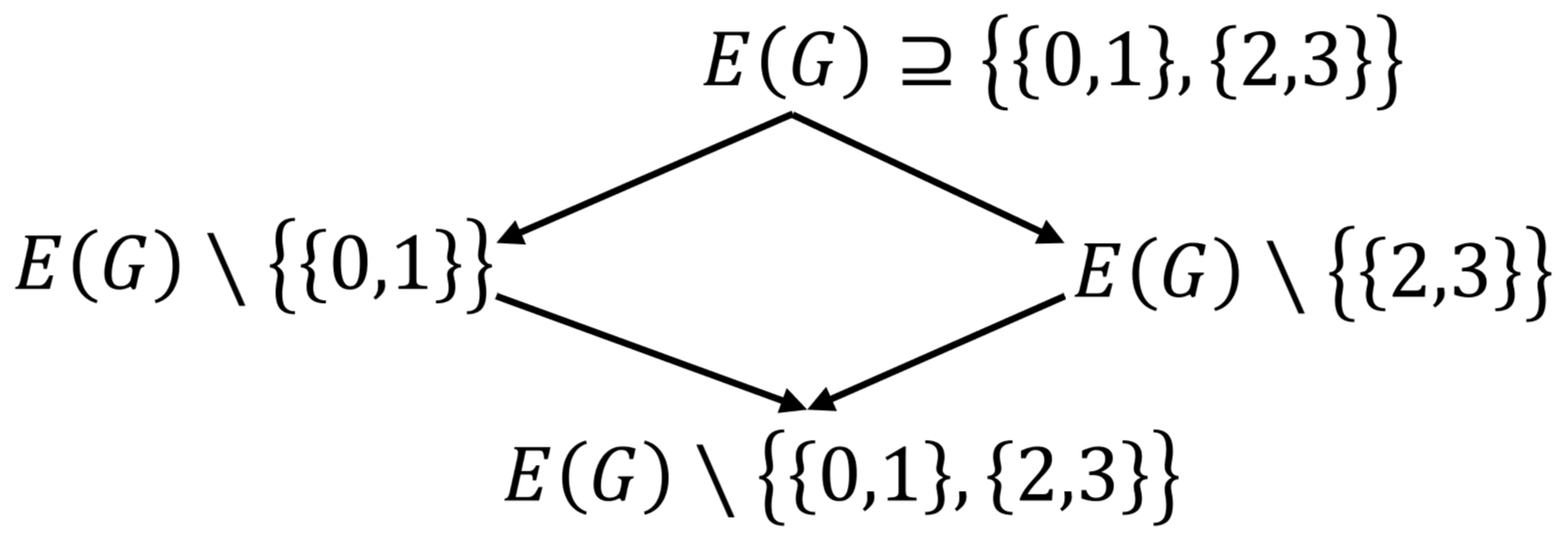}
	\caption{a lattice of subgraph of a given graph $G$ containing all interaction edges for $x_0x_1+x_2x_3+x_4+x_5$}
	\label{fig:subgraphLattice}
\end{figure}

The above figure shows the result in average.
However, the information about subgraphs is not shown in them, while it is another interesting property that can motivate us to construct an appropriate feature graph.
For the subgraph property, we consider the effect of removing the interaction edge from a graph.
For example, if we consider a graph $G$ containing all interaction edges, i.e. $\{\{0,1\},\{2,3\}\}\subseteq E(G)$ w.r.t. $x_0x_1 + x_2x_3 + x_4 + x_5$, the other considered subgraphs compared to this graph are $E(G)\setminus \{\{0,1\},\{2,3\}\}$, $E(G)\setminus \{\{0,1\}\}$ and $E(G)\setminus \{\{2,3\}\}$.
Figure \ref{fig:subgraphLattice} shows an abstract lattice of the subgraphs.
Our claim is that the performance would drop along this lattice from top to bottom of lattice.

To this end, we perform a statistical analysis by the confidence interval of the paired difference of MAE where the pair of graphs is paired with the subgraphs by removing the interaction edges: $\text{MAE}(G\setminus\{a,b\}) > \text{MAE}(G)$
where an edge $\{a,b\}$ is an interaction edge in a graph $G$.

\begin{figure}[h!]
	\centering
	\includegraphics[width=\linewidth]{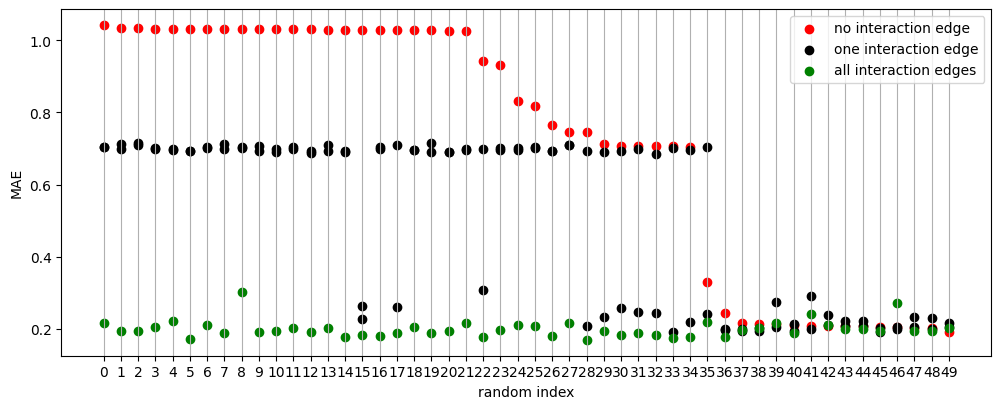}
	\caption{Samples of MAE score colored by the number of interaction edges and sorted by in x-axis by MAE of no-interaction-edge graphs to show the increasing of MAE when an interaction edge is removed, where our $G$ in this subsection corresponds to the green points}
	\label{fig:RemovingItrEdge}
\end{figure}

The lower-sided confidence interval 95\% is shown in Table \ref{tab:pval-remove-itrEdge}.
It ensures us that interaction edges are significantly important for input feature graph in attention-based GNNs.
Figure \ref{fig:RemovingItrEdge} shows the MAE of some random sample generated, sorted by the MAE of $E(G)\setminus \{\{0,1\},\{2,3\}\}$ from the 3 message passing layer model to illustrate how they are.

\begin{table}[h!]
	\centering
	\begin{tabular}{c c c c c }
		\hline
		graph   & dropped & 1MP & 2 MPs & 3 MPs \\ \hline
		$G$       & $\{0,1\}$  &  $>0.124$  &  $>0.173$  &  $>0.160$ \\
		$G$       & $\{2,3\}$  &  $>0.129$  &  $>0.186$  &  $>0.163$ \\ 
		$G\setminus\{0,1\}$ & $\{2,3\}$   &  $>0.114$  &  $>0.132$   &    $>0.137$ \\ 
		$G\setminus\{2,3\}$ & $\{0,1\}$  &  $>0.114$  &  $>0.156$   &  $>0.166$ \\ \hline
	\end{tabular}
	\caption{95\% lower-sided confidence interval of paired-difference of MAE between a graph and interaction-edge-removed graph}\label{tab:pval-remove-itrEdge}
\end{table}

\subsection{Reachability of Interaction Nodes and the Number of Message Passing Layers through a Longer Path}\label{subsec:reach}
\begin{figure*}[t!]
	\centering
	\includegraphics[width=\linewidth]{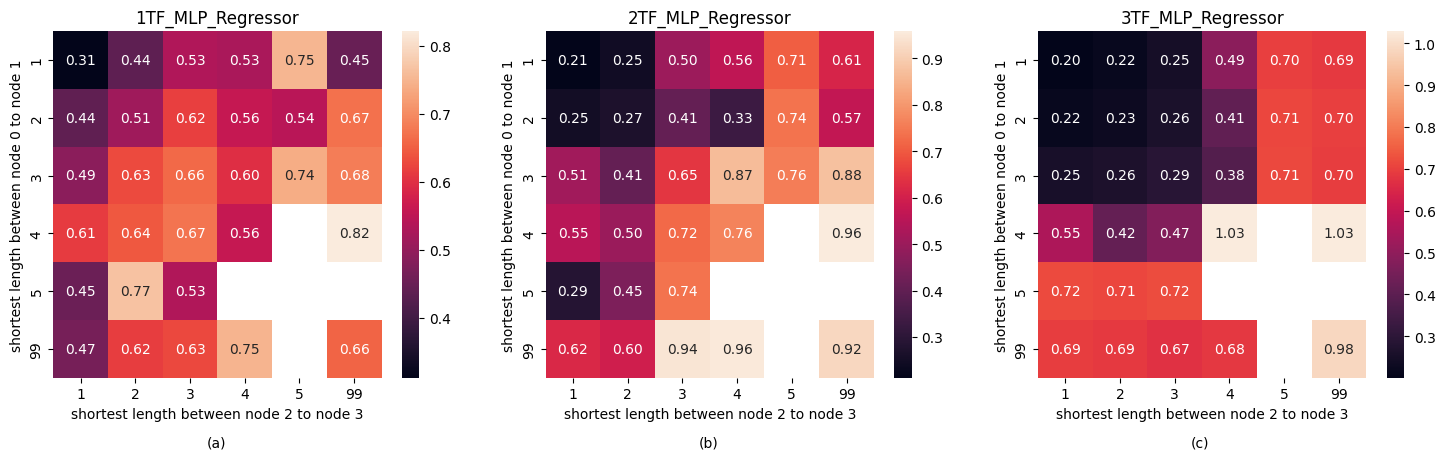}
	\caption{Heatmap of average MAE for the number of hops between interaction nodes (between node 0 and node 1 in vertical and between node 2 and node 3 in horizontal) for a model consists of (a) 1 MP layer, (b) 2 MP layers and (c) 3 MP layers: 99 means not reachable}
	\label{fig:result2}
\end{figure*}
In the previous result, we discussed the effect of the occurrence and absence of interaction and noninteraction edges.
We observed that the feature graphs without interaction edges can predict the test set better with the help of increasing non-interaction edges.
So, it comes up with the question about connectivity whether GNN models can capture interaction patterns via paths, not necessarily to be adjacent, and whether the length of a shortest path relates or not.

In Figure \ref{fig:result2} (a),
we see that although the interacting nodes are not adjacent, these graphs can still be used at some level if they are reachable.
In other words, connectivity through a sufficient number of hops can also yield satisfactory results.

We also see that the longer the shortest path between interacting nodes, the worse the score.
This result shows a considerable increase in error when the number of hops increases from 1 hop to 2 hops in both directions.
This might be because we used only one message passing layer in the model.
This allows the passing of messages between nodes only for one hop.

Since the passing message allows nodes to exchange their messages via edges, the greater number of layers should allow further nodes to do so.
So we show, in Figure \ref{fig:result2} (b)-(c), an alternative result experiment in which a GNN model consists of 2 and 3 message passing layers whose hyperparameters are set to have the same number of parameters as that of the 1-layer model.
The result is analyzed similarly to that shown in Figure \ref{fig:result2} (a).

In Figure \ref{fig:result2} (b) compared to (a), we see that the 2-layer model can be trained using a feature graph to ensure that the interaction nodes are reachable to each other up to 2 hops.
Similarly, Figure \ref{fig:result2} (c) of a 3-layer GNN reveals that the feature graph of up to 3 hops of interaction nodes can be used to train the GNN model.
This may infer that the more messages that pass through, the weaker the condition of reachability of the interaction node.

\subsection{Limitation of Complete Feature Graphs}\label{subsec:complete-graph}

From the result in Figure \ref{fig:result1-1}, the complete feature graph can give errors almost similar to that of the graph that contains only interacting edges.
This may sound like a good result to tell us that it is not necessary to find the optimal graph structures.
However, the demonstrating dataset is very small in number of features.
It contains only 6 features; then the complete graph consists of only 15 edges, which is small.
Therefore, we ask the question of how it will be when datasets contain more and more features.

\begin{figure}[h!]
	\centering
	a result from simulated datasets of equation $y=x_1x_2+\cdots + x_{2p-1}x_{2p} + x_{2p+1} + x_{2p+2}$
	\includegraphics[width=\linewidth]{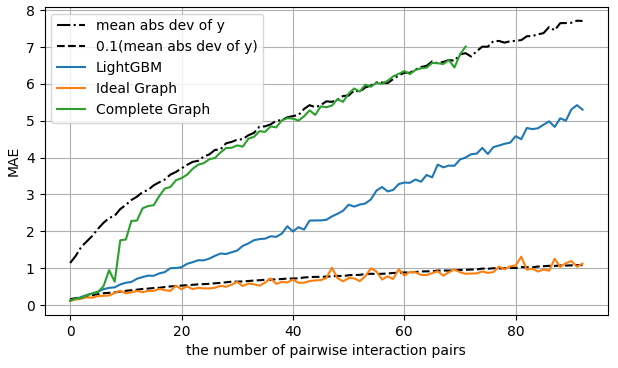}
	\caption{Predictive performance of lightGBM, GNN with complete graphs and GNN with ground truth feature graphs measured by MAE (y-axis) when the number of pairwise terms is varied (x-axis)}
	\label{fig:result3}
\end{figure}

Figure \ref{fig:result3} presents the results tested in synthetic datasets with 2 linear terms, that is, $y=x_1x_2+\cdots + x_{2p-1}x_{2p} + x_{2p+1} + x_{2p+2}$.
The results for complete graphs and the ground truth graphs are represented by the red and blue lines, respectively.
In addition, we depict the theoretical standard deviation of the datasets with a black dot line. The thick black line indicates the noise term $\epsilon$ added to the training label, which is $y + 0.1\sigma$.

It is evident that greater noise, resulting from an increase in the number of features, leads to poorer prediction performance.
When our GNN model is trained on complete feature graphs, it outperforms lightGBM when there are only a few pairwise terms (up to 10 in both datasets). Beyond this point, it dramatically increases, and the GNN model performs less effectively than lightGBM.

Notably, the red line plot in the right figure terminates at 40 pairwise terms (95 features or 4465 edges). The dataset in the right figure experiences memory leakage during the construction of complete feature graphs and computational processes of the model. This is why it is essential to avoid training the entire dataset using complete feature graphs when dealing with a larger number of features.

On the other hand, considering the performance of the GNN model trained on the ground truth feature graphs, the results show the lowest error compared to complete graphs and lightGBM. This underscores the value of pruning the edges of feature graphs when training models for prediction. Complete graphs not only lead to memory leaks but also capture uninformative hidden data. In conclusion, pairwise feature interaction is a beneficial property that should be retained in feature graphs to enhance the predictive performance of tabular data using GNNs.

\subsection{Discussion}
All results we show in this section mainly support the argument about the necessity of interaction edges in the case of a bilinear (pairwise interaction by multiplication) interaction.
Moreover, it also reveals that the non-interaction edges tend to behave like noise edges to capture non-underlying information in the given dataset when feature graphs contain all interaction edges.
So, if possible, we need to prune out such uninformative edges that avoid capturing noise in the dataset.
However, in some cases, we may not know explicit interaction pairs to be able to keep the correct edges.
Our result shows that the adjacency between interacting nodes can be weakened to only reachability via non-interaction edges in the limited number of hops by increasing the number of layers of message passing in the GNN models.

However, the results shown in this section are from only one simulated dataset.
Even though the results seem obviously to infer the conclusions, we need some theoretical results to guarantee the consistency of the results in any dataset that we are focusing on.
The results from Sections \ref{subsec:itr-vs-nonitr} and \ref{subsec:remove-itr} motivate us to reformulate the problem into a theoretical perspective which is discussed in the next section.

\section{Case Study: Click-through-rate Prediction}\label{sec:caseStudy}

We observed that it is ideal to include all known interaction edges in feature graphs. Non-interaction edges can be considered noise, capturing irrelevant information. When all interaction edges cannot be preserved, adjacency constraints can be relaxed to focus on reachability or connectedness, aided by more message passing layers.
In this section, we move from simulated data to the real world problems.
One of the real world problems that interactions between features are often concerned is the click-through rate (CTR) problem as a recommendation system in various application.
 
\begin{figure}[h!]
	\centering
	\includegraphics[width=1\linewidth]{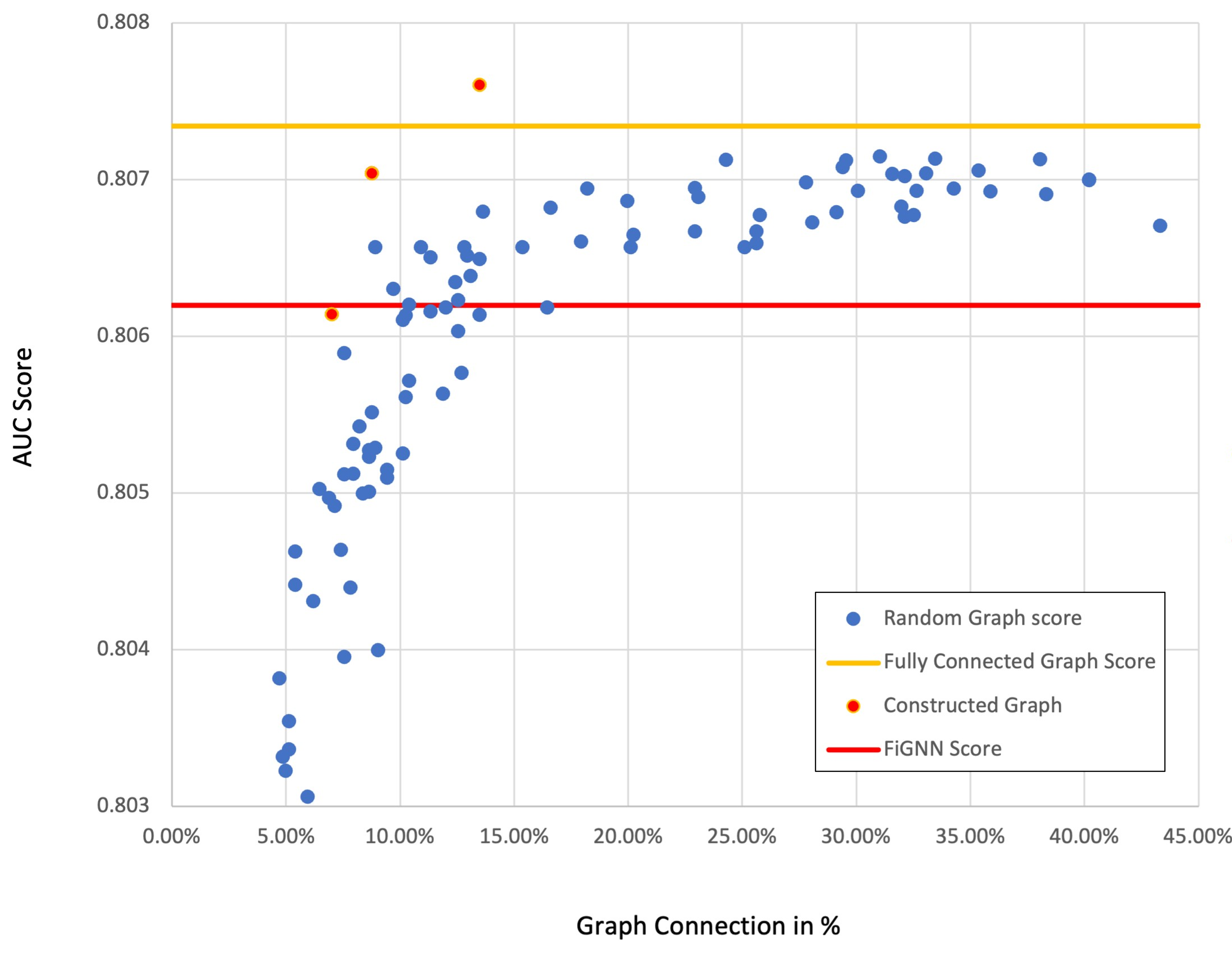}
	\caption{Comparison of AUC scores for different graph structures on the Criteo dataset. Randomly constructed sparse graphs (blue) exhibit varying performance based on connection percentage, with an optimal score of 0.8072 at 24\% connectivity. The manually constructed sparse graph (orange) outperforms both the fully connected graph (yellow) and FiGNN (red) with an AUC of 0.8076 at just 13\% connectivity.}
	\label{fig:resultctr}
\end{figure}

In this experiment, we utilized the Criteo dataset \cite{criteo2014} to evaluate the performance of different graph structures using the AUC score as the primary metric. We adopt FiGNN, a widely recognized graph model for CTR prediction, as our baseline. FiGNN utilizes the complete graph and learns the dynamic graph structure for each sample, achieving an AUC score of 0.8062. 

Initially, we generated random sparse graphs with varying percentages of connections and evaluated their performance using a simple GNN model. The optimal AUC score for randomly constructed graphs was approximately 0.8072, achieved with 24\% graph connections.
Subsequently, we experimented with a fully connected graph, utilizing 100\% of the graph connections. The AUC score for the fully connected graph was 0.8073.
Additionally, we manually constructed a sparse graph inspired by the GDCN heatmap \cite{GDCN}. This manually constructed graph achieved an AUC score of 0.8076 with only 13 \% of the connections.

The findings of this experiment demonstrate that a simple GNN model with a randomly constructed sparse graph can outperform the FiGNN model and achieve results comparable to a fully connected graph. Moreover, the manually constructed sparse graph, designed with connections that align with the feature interactions, outperforms both the fully connected graph and FiGNN in performance.

\section{Conclusion and Future Work} \label{sec:discussion}

In this study, we investigated the relationship between the structure of feature graphs and their capacity to represent pairwise feature interactions within GNNs. We offered both theoretical and empirical insights on the development of sparse feature graphs. Our findings reveal that interaction edges are crucial, whereas non-interaction edges can introduce noise, which hinders the learning process of GNN models. From a theoretical standpoint, employing the Minimum Description Length (MDL) principle, we showed that feature graphs that retain only essential interaction edges provide a more efficient and interpretable representation compared to complete graphs. Empirical tests on synthetic datasets underscored the significance of interaction edges for improved prediction performance, while non-interaction edges were found to introduce noise that reduces model accuracy. Additionally, we discovered that the connection between interacting nodes, even without direct interaction edges, can enhance learning via multi-hop message passing in GNNs.

Our findings emphasize the significance of feature graph sparsity for both computational efficiency and prediction accuracy, particularly in large-scale datasets where complete graphs become useless.
Furthermore, we validated our insights through a case study on click-through rate prediction, further highlighting the practical value of constructing well-designed feature graphs for real-world tasks.

For future work, our aim is to extend this study by:
\begin{enumerate}
	\item Investigating algorithm for constructing optimal feature graphs from raw datasets, especially when prior knowledge about feature interactions is unknown.
	\item Exploring feature interactions beyond pairwise relationships, such as higher-order interactions, and their implications for feature graph design.
	\item Integrating dynamic graph construction methods to adapt feature graphs during model training for each instance, allowing better handling of evolving datasets and relationships.
	\item Design new message-passing mechanisms specifically tailored for interaction modeling, enabling GNNs better to capture interaction patterns and relationships in feature graphs.
\end{enumerate}

By addressing these directions, we aim to enhance the understanding and utility of feature graphs in GNN-based models, thereby contributing to improved model performance and explainability across a wide range of applications.

\section*{Acknowledgment}
Anonymous submission.


\bibliographystyle{IEEEtran}

\end{document}